\documentclass[letterpaper, 10 pt, conference]{ieeeconf}  

\usepackage{graphics} 
\usepackage{amsmath} 
\usepackage{amssymb}  
\usepackage[]{algorithm}
\usepackage{algorithmic}
\usepackage{url}
\usepackage{array}
\usepackage{amsfonts}
\usepackage{caption}
\usepackage{mathtools}
\usepackage{bm} 
\usepackage{tabularx}
\usepackage{epstopdf} 
\usepackage{multirow,booktabs} 
\usepackage{alltt}
\usepackage{bigstrut}
\usepackage{tabularx,ragged2e,booktabs}
\usepackage{subcaption}
\usepackage{balance}

\usepackage[utf8]{inputenc}
\def \l {{\boldsymbol{\ell}}}
\def \r {{\mathbf{r}}}
\def \k {{\mathbf{k}}}
\def \s {{\textbf{s}}}
\def \p {\textbf{\emph{p}}}

\def \f {\boldsymbol{f}}
\def \fb {\mathbf{f}}
\def \g {\boldsymbol{g}}
\def \q {\textbf{q}}
\def \Z {\mathbf{Z}}

\def \e {{\mathbf{e}}}
\def \U {{\mathcal{U}}}
\def \D {{\mathcal{D}}}
\def \V {{\mathcal{V}}}
\def \db {{\boldsymbol{\delta}}}

\def \Z {{\mathcal{Z}}}
\def \E {{\mathcal{E}}}
\def \W {{\mathbf{W}}}
\def \mub {{\boldsymbol{\mu}}}
\def \K {{\mathbf{K}}}
\def \Rn {{\mathbb{R}}}
\def \Sig {{\boldsymbol{\Sigma}}}
\def \O {{\mathcal{O}}}
\def \x {{\boldsymbol{x}}}
\makeatletter
\newcommand*\bigcdot{\mathpalette\bigcdot@{1}}
\newcommand*\bigcdot@[2]{\mathbin{\vcenter{\hbox{\scalebox{#2}{$\m@th#1\bullet$}}}}}

\makeatother

\title{\LARGE \bf
Model Free Calibration of Wheeled Robots Using Gaussian Process   
}

\author{Mohan Krishna Nutalapati,~\IEEEmembership{Student Member,~IEEE}
Lavish Arora,~\IEEEmembership{Student Member,~IEEE} Anway Bose, Ketan Rajawat,~\IEEEmembership{Member,~IEEE} and 
Rajesh M Hegde~\IEEEmembership{Senior Member,~IEEE} 
\thanks{{The authors are with the Department of Electrical
		Engineering, Indian Institute of Technology Kanpur, Kanpur 208016, India,
		(e-mail: $\left\{\text{nmohank, lavi, anwayb, ketan, rhegde} \right\}$@iitk.ac.in).}}}

\begin{document}

\maketitle
\thispagestyle{empty}
\pagestyle{empty}

\begin{abstract}
Robotic calibration allows for the fusion of data from multiple sensors such as odometers, cameras, etc., by providing appropriate relationships between the corresponding reference frames. For wheeled robots equipped with camera/lidar along with wheel encoders, calibration entails learning the motion model of the sensor or the robot in terms of the data from the encoders and generally carried out before performing tasks such as simultaneous localization and mapping (SLAM). 
This work puts forward a novel Gaussian Process-based non-parametric approach for calibrating wheeled robots with arbitrary or unknown drive configurations. The procedure is more general as it learns the entire sensor/robot motion model in terms of odometry measurements. Different from existing non-parametric approaches, our method relies on measurements from the onboard sensors and hence does not require the ground truth information from external motion capture systems. Alternatively, we propose a computationally efficient approach that relies on the linear approximation of the sensor motion model. Finally, we perform experiments to calibrate robots with un-modelled effects to demonstrate the accuracy, usefulness, and flexibility of the proposed approach.

\end{abstract}

\section{INTRODUCTION}
{R}{obotic} calibration is an essential first step necessary for carrying out various sophisticated tasks such as simultaneous localization and mapping (SLAM) \cite{SLAM1, SLAM2}, object detection and tracking \cite{objectdetec}, and autonomous navigation \cite{Autonav}. For most wheeled robot configurations equipped with wheel encoders and exteroceptive sensors like camera/lidar, the calibration process entails learning a mathematical model that can be used to fuse odometry and sensor data. In the case when the motion model of the robot is unavailable, due to some unmodelled effects calibration involves learning the relationships that describe the sensor motion in terms of the odometry measurements. Precise calibration is imperative since calibration errors are often systematic and tend to accumulate over time \cite{new}. Conversely, an accurately specified odometric model complements the exteroceptive sensor, e.g. to correct for measurement distortions if any \cite{LOAM}, and continues to provide motion information even in featureless or geometrically degenerate environments \cite{degraded}.

{Traditional approaches} \cite{extend,intrinsic3} for calibration of wheeled robots focus on learning a parametric motion model of the robot/sensor. A common issue among these approaches was the need for external measurement setup such as calibrated video cameras or motion capture systems. On the other hand \cite{censi,tricycle} overcome this issue by performing simultaneous calibration of odometry and sensor parameters using measurements from the sensor. More generic calibration routines for arbitrary robot configurations were presented in \cite{agv25,SLAMC} where solution to calibration parameters is found along with robot state variables. All these techniques essentially learn the parameters associated with the motion model of the robot/sensor to generate accurate odometry. However, the performance degrades due to uncertainty arising from interactions with the ground and hence lead to bad odometry estimates \cite{new}. Moreover, modeling such uncertainties that arise due to non-systematic errors is a very challenging and difficult task. To this end, non-parametric methods \cite{blimp,new} employ tools from Gaussian process (GP) estimation learn the residuals between the parametric model and ground truth measurements from external motion capture systems. A common assumption that all these methods make is the residual function being zero mean, which holds true only when the kinematic model of the robot is accurately known. However, for robots with misaligned wheel axis or other unknown offsets that may arise due to unsupervised assembly \cite{nomodel}, excessive wear-and-tear etc., zero mean assumption is not valid rendering the approach suboptimal.

\begin{figure}
\begin{subfigure}{.25\textwidth}
  \centering
  \includegraphics[width=.90\linewidth]{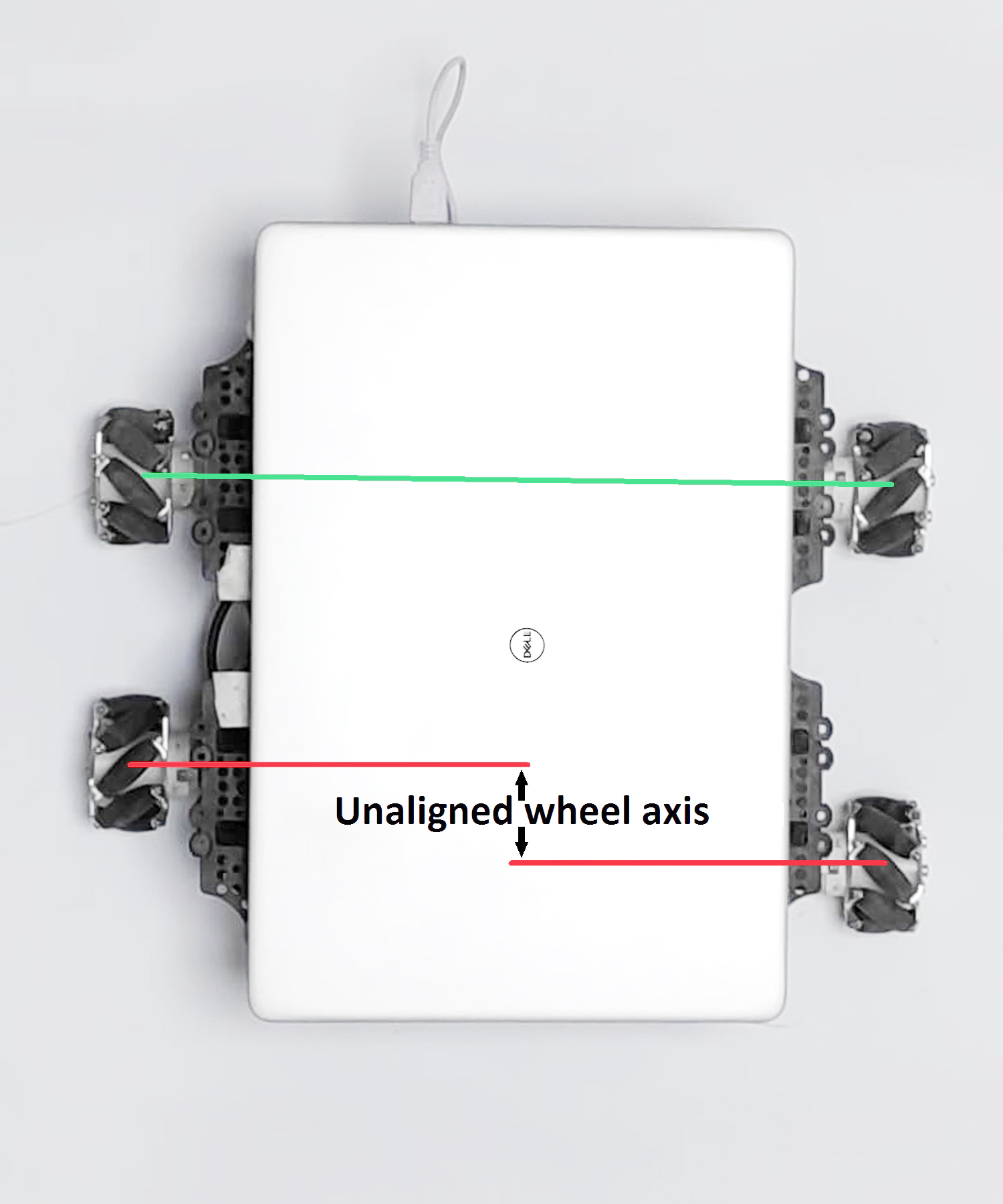}
  \caption{Configuration \textbf{T1}} 
\end{subfigure}%
\begin{subfigure}{.25\textwidth}
  \centering
  \includegraphics[width=.90\linewidth]{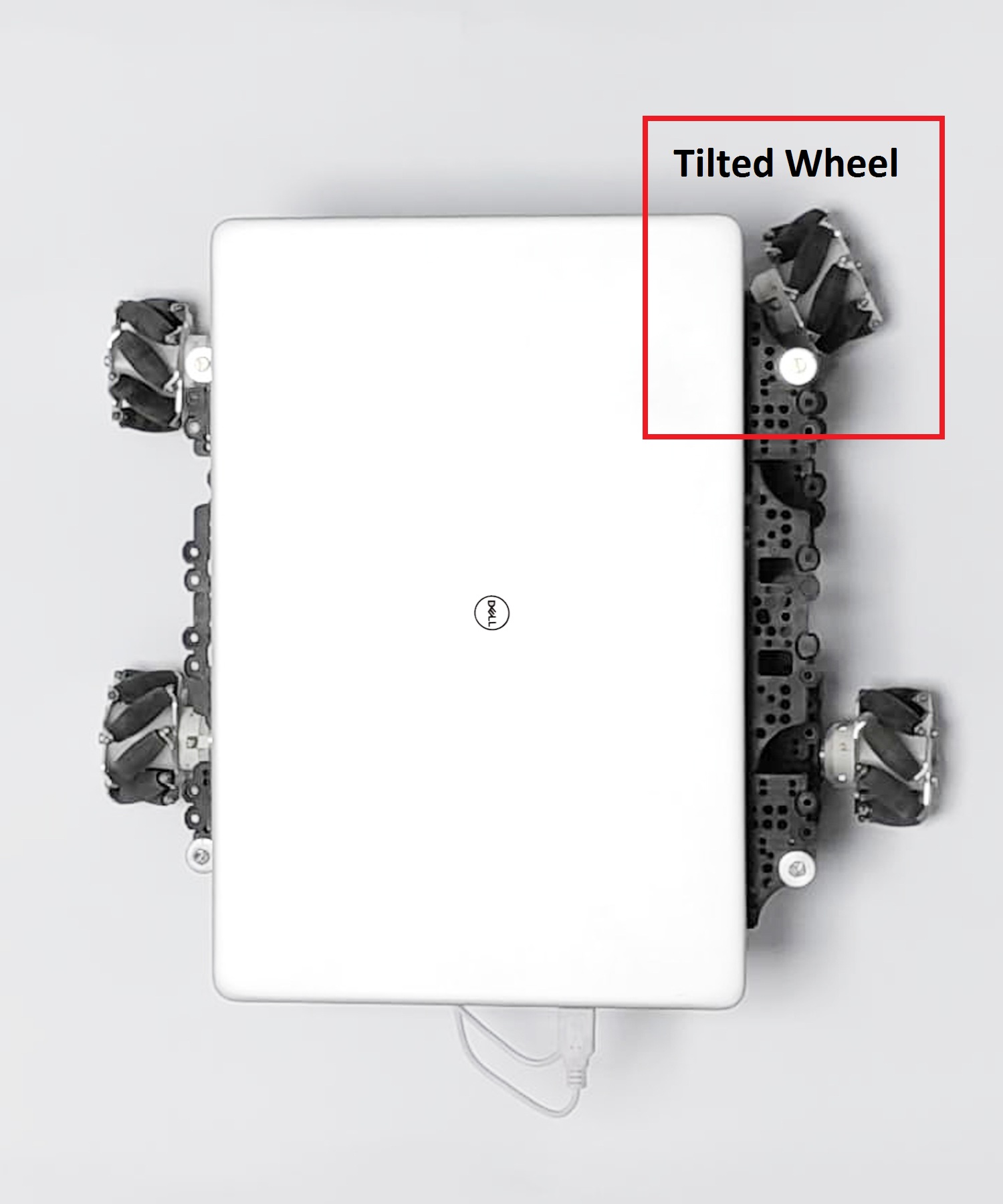}
  \caption{Configuration \textbf{T2}} 
\end{subfigure}
\caption{Deformed Turtlebot3 mecanum drive robot used for experimental evaluations. (a) Unaligned wheel axis deformation, (b) Tilted wheel deformation.}\label{meca}
 \vspace{-0.6cm}
\end{figure}

This work puts forth a more general framework that subsumes existing non-parametric approaches, while also applicable to scenarios where the motion model of the robot/sensor is distorted or not known. Different from the existing non-parametric approaches, the proposed method learns the whole sensor/robot motion model. To this end, the key contributions of the work are
\begin{itemize}
\item Formulation of a Gaussian-process regression framework that captures the arbitrary or unknown motion model of the sensor/robot. The entire calibration routine can be carried out using measurements from onboard sensors capable of sensing ego-motion  
\item A computationally efficient approach for near-optimal and model-free calibration  
\end{itemize}   
The rest of the paper is organized as follows. Sec. II details the system setup and the problem formulation. The proposed algorithm is described in Sec. III. Detailed experimental evaluations are carried out to validate the performance of the proposed method, and the results are discussed in Sec. IV. Finally, Sec. V concludes the paper. The notation used in the paper is summarized in Table \ref{notation}. 

\section{Problem Formulation} \label{PF}

\begin{table}
 \fontsize{8pt}{8pt}\selectfont
\centering
\caption{Nomenclature used in the paper}
\label{notation}
\begin{tabular}{@{}l@{}}
\toprule
Parameters 
\\ \midrule
$\begin{matrix} \p = (\underbrace{\ell_x,\ell_y,\ell_\theta}_\l,\r) &  \text{parameters to be estimated} \end{matrix}$ \\
$\begin{matrix} \l  &  \text{position of extrinsic sensor w.r.t robot frame} \end{matrix}$ \\
$\begin{matrix} \r  &  \text{robot instrinsic parameters} \end{matrix}$ \\
\midrule
Measurements \\ \midrule
$\begin{matrix}
\mathcal{U}  & \text{raw data  log of odometry sensor} & \\
\mathcal{V}  & \text{Measurememts from exteroceptive sensor} & \\ 
\q(t)\ \   = &[q_x(t)\ q_y(t)\ q_\theta(t)]^T\ \text{Pose of robot at any time }\emph{t} \\ 
\hat{\s}_{jk}  & \text{sensor displacement estimate for time interval}\ \emph{$[t_j,t_k)$} 
\end{matrix}$ \\
\midrule 
More Symbols \\
\midrule
$\begin{matrix}
\oplus &\text{Roto-translation operator}  & \\ 
\circleddash &\text{inverse of} \oplus \text{operator}   & \\
 & \begin{bmatrix}
a_x\\ 
a_y\\ 
a_\theta
\end{bmatrix} \oplus \begin{bmatrix}
b_x\\ 
b_y\\ 
b_\theta
\end{bmatrix} \overset{\Delta}{=} \begin{bmatrix}
a_x + b_x \cos a_\theta - b_y \sin a_\theta\\ 
a_y + b_x \sin a_\theta + b_y \cos a_\theta \\ 
a_\theta + b_\theta
\end{bmatrix} & \\ \\ 
& \circleddash \begin{bmatrix}
a_x\\ 
a_y\\ 
a_\theta
\end{bmatrix} \overset{\Delta}{=} \begin{bmatrix}
-a_x \cos a_\theta - a_y \sin a_\theta\\ 
a_x \sin a_\theta - a_y \cos a_\theta\\ 
-a_\theta
\end{bmatrix}

\end{matrix}$ \\
\midrule
 
\end{tabular}

\vspace{-0.5cm}
\end{table}

In this section we first start with introducing preliminary notations (see Table \ref{notation}) used through out the paper. Consider a general robot with an arbitrary drive configuration, equipped with $m$ rotary encoders on its wheels and/or joints and an exteroceptive sensor such as a lidar or a camera. The exteroceptive sensor can sense the environment and generate scans or images $\V=\{\Z(t)\}_{t\in \mathcal{T}}$ that can be used to estimate its ego motion. Here, $\mathcal{T}:=\{t_1, t_2, \ldots, t_n\}$ denotes the set of discrete time instants at which the measurements are made. The rotary encoders output raw odometry data in the form of a sequence of wheels angular velocities $\U = \{\db(t)\}_{t\in \mathcal{T}}$. Given two time instants $t_j$ and $t_k$ such that $\Delta t_{jk}:=t_k-t_j > 0$ is sufficiently small, it is generally assumed that $\db(t):=\db_{jk}$ for all $t_j \leq t < t_k$. Traditionally, the odometry data is pre-processed to yield relative translation motion and orientation information, and is subsequently fused with the ego motion estimates from exteroceptive sensors. This pre-processing step necessitates the use of the motion model $\f_r$ of the robot that acts upon the odometry data $\db_{jk}$ to yield the relative pose of the robot $\q_{jk} :=  \circleddash \q_j \oplus \q_k =\f_r(\db_{jk})$ for the time interval $\Delta t_{jk}$. Here $\q_j :=  (q_j^x,q_j^y,q_j^\theta)^\intercal$ denote the position of the robot at time $t = t_j$. Note that if the exteroceptive sensor is mounted exactly on the robot frame of reference, the sensor motion model denoted by $\f$ is the same as the robot motion model $\f_r$. In general however, if the pose of the exteroceptive sensor with respect to the robot is denoted by $\l$, the sensor motion model is given by $\f(\db_{jk}) = \circleddash \l \oplus \f_r(\db_{jk}) \oplus \l$, where generally $\l$ is unknown. 
\begin{figure} 
      \begin{subfigure}{.48\textwidth}
         \centering 
              \includegraphics[width=0.99\linewidth]{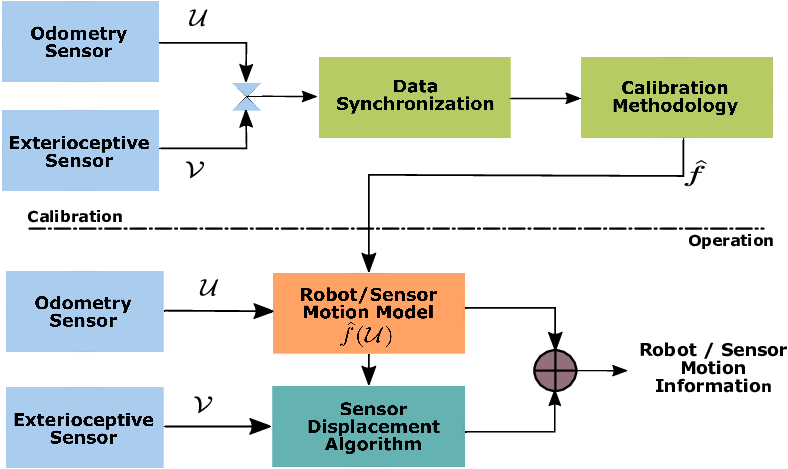}
      \end{subfigure}%
       \caption{Block diagram describing the system setup and data flow }\label{meth}  \vspace{-0.5cm}
\end{figure}

Having the preliminary notations at hand, we now describe the system setup displayed in Fig. \ref{meth}. The goal of the calibration phase is to estimate the function $\f$, given $\mathcal{U}$ and $\mathcal{V}$. The estimated motion model, denoted by $\hat{\f}$, is subsequently used in the operational phase to augment or even complement the motion estimates provided by the exteroceptive sensor. For instance, accurate odometry can be used to correct distortions in the sensor measurements \cite{LOAM}. Note here that the parametric form of the function $\f$ exists when the robot motion model $\f_r$ is well defined. For instance two-wheel differential drive robot \cite{censi} , four wheel mecanum drive \cite{mecanum} etc. In other words,  $\f(\bigcdot) = \g(\bigcdot\ ;\ \p)$ where $\g$ is a known function and $\p$ is the set of unknown parameters, such as the dimensions of the wheel, sensor position w.r.t robot frame of reference etc. 
State-of-the-art techniques like \cite{censi,tricycle,trad1} learns $\f$ under this assumption. A significantly more challenging scenario occurs when the form of $\f$ is not known, e.g. due to excess wear-and-tear, or is difficult to handle, e.g. due to non-differentiability. For such cases, the parametric approaches \cite{censi,tricycle,trad1} are no longer feasible and the unknown function $\f$ is generally infinite-dimensional. Towards this end, a low-complexity approach is proposed (see Sec. \ref{linear_model}), wherein a simple but generic (e.g. linear) model for $\f$ is postulated. A more general and fully non-parametric Gaussian process framework is also put forth that is capable of handling more complex scenarios and estimate a broader class of motion models $\f$. It is remarked that in this case, unless the exteroceptive sensor is mounted on the robot axis, additional information may be required to also estimate the robot motion model $\f_r$.

\section{Model Free Calibration using GP} \label{GP}
When no information about the kinematic model of the robot is available, it becomes necessary to estimate $\f$ directly. As in Sec. \ref{PF}, let $\mathcal{T}$ be the set of time instants at which measurements are made. For certain time interval $[t_j,t_k)$ for which $\Delta t_{jk}$ is not too large, let the exteroceptive sensor generates motion estimates $\hat{\s}_{jk}$. Given data of the form $\D:=(\db_{jk}, \hat{\s}_{jk})_{(j,k)\in\E}$, where $\E$ represents set of indices of all chosen key measurement pairs of the sensor and $n:=|\E|$, the goal is to learn the function $\f:\Rn^m\rightarrow \Rn^3$ that adheres to the model
\begin{align}
\hat{\s}_{jk} = \f(\db_{jk}) + \boldsymbol{\varepsilon}_{jk}
\end{align}
for all $(j,k)\in\E$, where $\boldsymbol{\varepsilon}_{jk} \sim  \mathcal{N}(\boldsymbol{\mu},\boldsymbol{\Sigma}_{jk}) \in \Rn^{3}$ models the noise in the measurements, and  $\db_{jk}$ now represents the number of wheel ticks recorded in the time interval $\Delta t_{jk}$. Here we assume that the noise covariance $\Sig_{jk}$ is known before hand. Given an estimated $\hat{\f}$ of the sensor motion model, a new odometry measurements $\db_e$ for an edge $e$ (pair of times at which measurements are made) can be used to directly yield sensor pose changes $\s_e = \hat{\f}(\db_e)$. As remarked earlier, it may be possible to obtain the robot pose change $\q_e$ from $\s_e$ if the sensor pose $\l$ is known a priori. Since the functional variable $\f$ is infinite dimensional in general, it is necessary to postulate a finite dimensional model that is computationally tractable. Towards solving the functional estimation problem, we detail two methods, that are very different in terms of computational complexity and usage flexibility.


\begin{figure*}
\begin{subfigure}{.33\textwidth}
  \centering
  \includegraphics[width=0.90\linewidth]{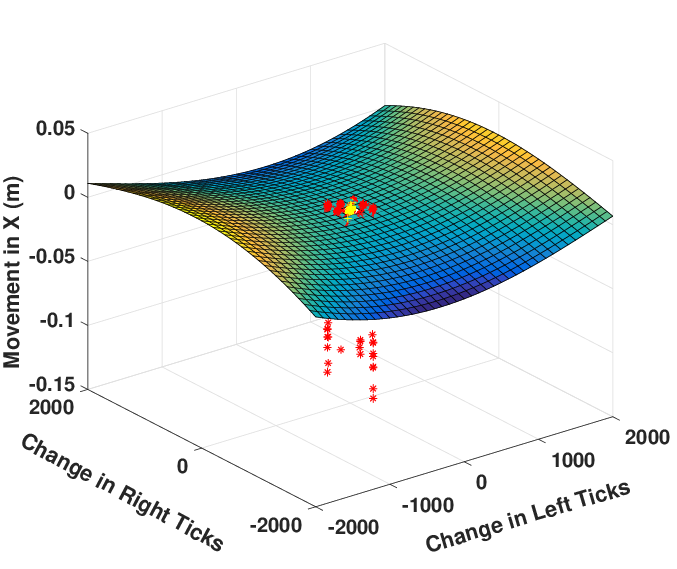}
  \caption{x} 
\end{subfigure}%
\begin{subfigure}{.33\textwidth}
  \centering
  \includegraphics[width=0.90\linewidth]{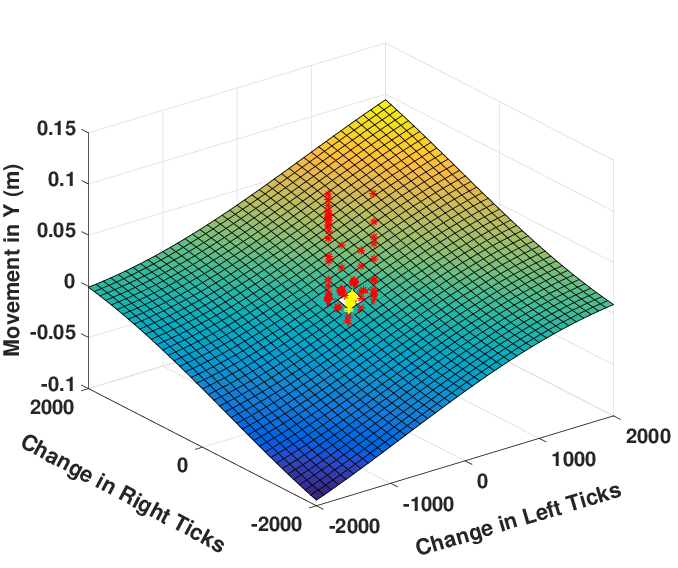}
  \caption{y} 
\end{subfigure}
\begin{subfigure}{.33\textwidth}
  \centering
  \includegraphics[width=0.90\linewidth]{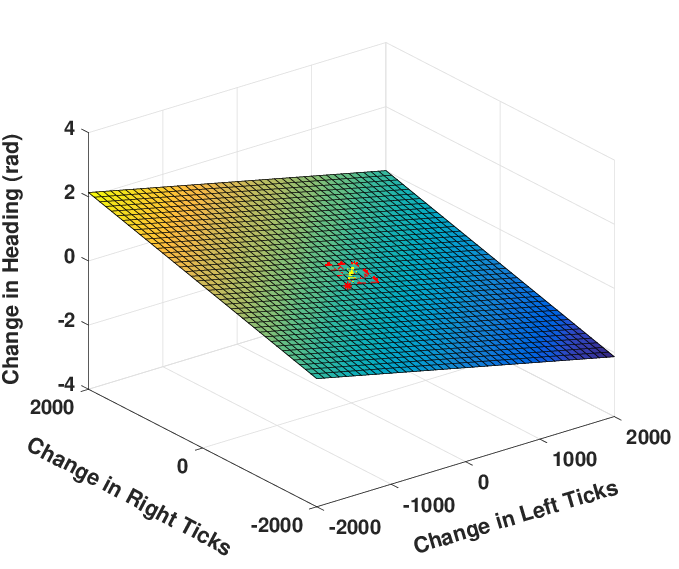}
  \caption{$\theta$} 
\end{subfigure}%
\\
\begin{subfigure}{.33\textwidth}
  \centering
  \includegraphics[width=0.90\linewidth]{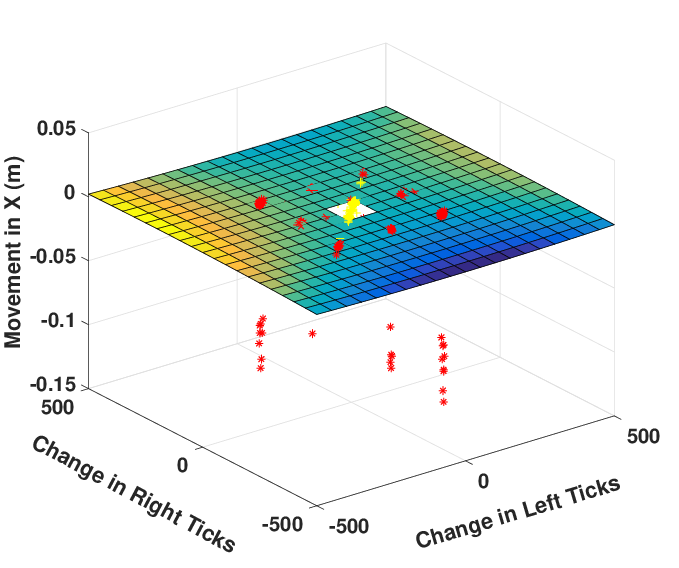}
  \caption{x} 
\end{subfigure}
\begin{subfigure}{.33\textwidth}
  \centering
  \includegraphics[width=0.90\linewidth]{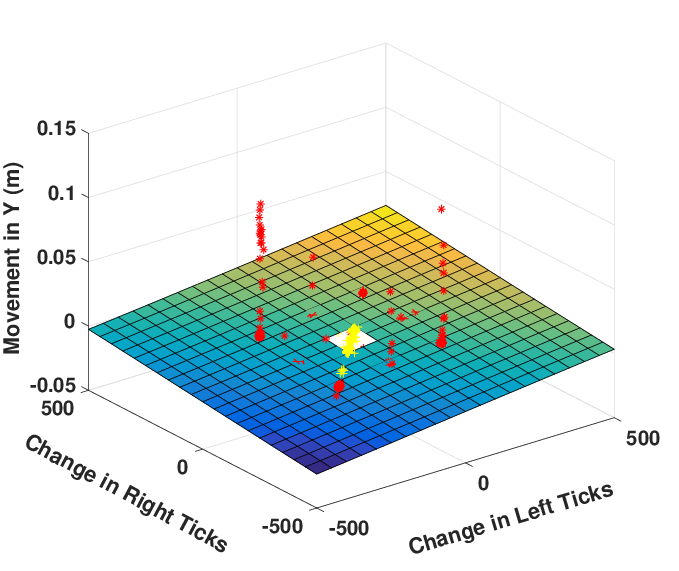}
  \caption{y} 
\end{subfigure}
\begin{subfigure}{.33\textwidth}
  \centering
  \includegraphics[width=0.90\linewidth]{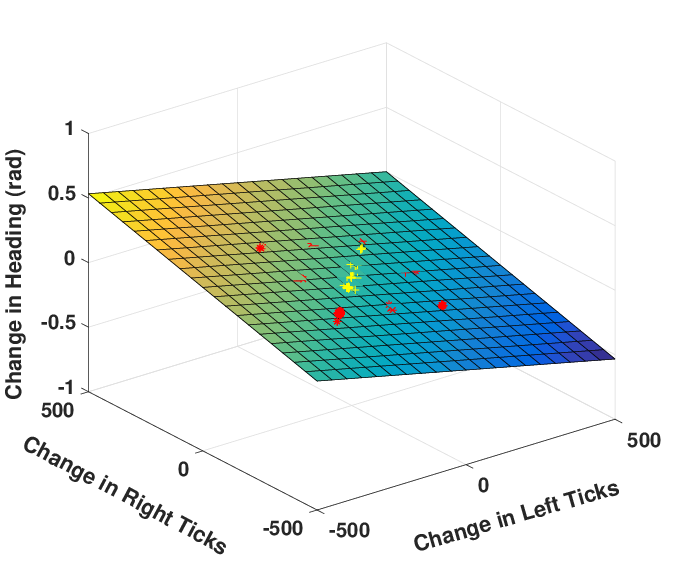}
  \caption{$\theta$} 
\end{subfigure}
\caption{ (a),(b),(c) illustrates the movement of the sensor frame in $x,y,\theta$, respectively w.r.t change in left and right wheel ticks of a two wheel differential drive robot (i.e., $\f(\bullet) = \g(\bullet \ ;\ \hat{\p})$), overlaid with the corresponding sensor displacement measurements generated using raw data published at \cite{web} for a particular configuration. Note $\hat{\p}$ denotes parameter estimates found using CMLE \cite{censi}. Also in \cite{web}, since data from each configuration is divided into three subsets, we consider any two of them as training data and rest as test data. Points that are displayed in red and yellow color denote training and testing samples generated at selected scan instants respectively. Note: red points that are away from the 3D surface are outliers. (d)-(f) represent the truncated and enlarged versions of the same plots to expose the linearity. (Figure is best viewed in color)} \label{fig2}
\vspace{-0.2cm}
\end{figure*}

\subsection{Calibration via Gaussian process regression (CGP)}
The GP regression approach assumes that the measurement is Gaussian distributed and that the function $\f$ is  a Gaussian process, whose mean and variance functions depend on the data. Specifically, we have that 
\begin{align}
\hat{\s}_{jk} \sim \mathcal{N}(\f(\db_{jk}), \Sig_{jk})
\end{align}
or equivalently, $\boldsymbol{\varepsilon}_{jk} \sim \mathcal{N}(\textbf{0},\Sig_{jk})$. Given inputs $\{\db_{jk}\}$, let $\fb$ denote the $\{3n \times 1\}$ vector that collects $\{\f(\db_{jk})\}$ for $\{(j,k)\in\E \}$. Defining $\Sig \in \Rn^{3n \times 3n}$ as the block diagonal matrix with entries $ \Sig_{jk}$ and $\hat{\s} \in \Rn^{3n}$ as the vector that collects all the measurements $\{\hat{\s}_{jk}\}_{(j,k)\in\E}$. Having this we can equivalently write the joint likelihood as
\begin{align}
p(\hat{\s}|\fb) = \mathcal{N}(\hat{\s}|\fb, \Sig)
\end{align}

Unlike the parametric model based approaches \cite{censi}, we impose a Gaussian process prior on $\f$ directly. Equivalently, we have that
\begin{align}
p(\fb) = \mathcal{N}(\fb|\mathbf{\bar{\mub}}, \K)
\end{align}
where $\bar{\mub} \in \Rn^{3n}$ is the mean vector with stacked entries of $\mub(\db_{jk}) \in \Rn^3$, and $\K \in \Rn^{3n \times 3n}$ is the covariance matrix with a block of entries $[\K_{i,i'}] = \boldsymbol{\kappa}(\db_{jk},\db_{j'k'})$ for $(j,k)$ and $(j',k') \in \E$ and $i,i' \in \{1,\cdots, n\}$. The choice of the mean function $\mub:\Rn^m \rightarrow \Rn^3 $ and the kernel function $\boldsymbol{\kappa} :\Rn^{m} \times \Rn^m \rightarrow \Rn^{3\times3}$ is generally important and application specific. Popular choices include the linear, squared exponential, polynomial, Laplace, and Gaussian, among others. With a Gaussian prior and noise model, the posterior distribution of $\f$ given $\D$ is also Gaussian. For a new odometry measurement $\db_e$ with noise variance $\Sig_e$, let $\k_e \in \Rn^{3n \times 3}$ be the vector that collects $\{\boldsymbol{\kappa}(\db_e, \db_{jk})\}_{(j,k)\in\E}$. Then the distribution of $\hat{\f}(\db_e)$ for given $\hat{\s}$ is
\begin{align}
p(\hat{\f}(\db_e)|\hat{\s}) = \mathcal{N}(\hat{\f}(\db_e)\ | \ \hat{\mub}_e, \hat{\Sig}_e)
\end{align}
where $\hat{\mub}_e = \k_e^T(\K+\Sig)^{-1}(\hat{\s}-\bar{\mub}) + \mub(\db_e) $ and the covariance $\hat{\Sig}_e = \boldsymbol{\kappa}(\db_e,\db_e) - \k_e^T(\K+\Sig)^{-1}\k_e$.
Note that in general, the choice of the mean and kernel functions is important and specific to the type of robot in use. In the present case, we use the linear mean function   
	\begin{equation}
	\mub(\textbf{x}) = \textbf{C} \textbf{x}
	\end{equation}
where $\textbf{x} \in \Rn^m $ is the vector of wheel ticks recorded in a time interval and $\textbf{C} \in \Rn^{3\times m}$ is the associated hyper-parameter of the mean function. Recall that $m$ represents total number of wheels equipped with wheel encoders. Intuitively, the implication of this choice of linear mean function is that the relative position of the robot varies linearly with the wheel ticks recorded in the corresponding time interval. Such a relationship generally holds for arbitrary drive configurations if the time interval is sufficiently small.
A widely used kernel function is the radial basis function as follows 
\begin{equation}\label{covse}
[\boldsymbol{\kappa}_{rbf}(\textbf{x},\textbf{x}')]_{i,i'} = \sigma_{i,i'}^2 \exp\left(-\frac{1}{2}(\textbf{x}-\textbf{x}')^T \textbf{B}_{i,i'}^{-1}(\textbf{x}-\textbf{x}')\right)
\end{equation}
where $\textbf{x},\textbf{x}'$ $\in \mathbb{R}^m$ are the data inputs with hyper-parameters $\Xi = [\sigma_{i,i'},\textbf{B}_{i,i'}]$, here $i,i' = 1,2,3$. It will be shown in section \ref{expres} that for the two-wheel differential drive robot in use here, the squared exponential kernel \eqref{covse} with the linear mean function yielded better results than others. On the other hand for four-wheel Mecanum drive in use here, the inner product kernel, which amounts to a linear transformation of the feature space, 
\begin{equation}
[\boldsymbol{\kappa}_{lin}(\textbf{x},\textbf{x}')]_{i,i'} = \langle\ \textbf{x},\textbf{x}'\rangle
\end{equation}
performed better. We remark here that for our experiments we have assumed $\boldsymbol{\kappa}(\textbf{x},\textbf{x}')$ is a diagonal matrix with diagonal entries $\{[\boldsymbol{\kappa}(\textbf{x},\textbf{x}')]_{i,i}\}$. In general, the choice of the mean and kernel functions and that of the associated hyper-parameters is made a priori. For our experiments we infer the hyper-parameters by optimizing the corresponding log marginal likelihood. However, they may also be determined during the calibration phase via cross-validation.  

\begin{algorithm} 
    	\caption{CGP algorithm to learn the motion model of the sensor} 	\label{algo3}
    	 \begin{algorithmic}[1]
           \STATE  Collect measurements from sensors. 
           \STATE \textbf{Training Phase :}
           \STATE  Run sensor displacement algorithm for each selected interval, to get the estimates  $\{\hat{\s}_{jk}\}$ with the corresponding wheel ticks $\db_{jk}$ and stack them.
           	\STATE Now pre-compute the following quantities :  
           		\STATE \hspace{1cm}$(\boldsymbol{K}+\boldsymbol{\Sigma})^{-1}(\hat{\s}-\bar{\mub}) $ and $(\boldsymbol{K}+\boldsymbol{\Sigma})^{-1}$    
           		\STATE \textbf{Testing Phase :}
           		\STATE For every test input $\db_e$, evaluate the following,    
           		\STATE \hspace{1cm} $\hat{\mub}_{e}  = \boldsymbol{k}_e(\boldsymbol{K}+\boldsymbol{\Sigma})^{-1}(\hat{\s}-\bar{\mub}) + \mub(\db_e)$ 
           		\STATE \hspace{1cm} $\hat{\boldsymbol{\Sigma}}_e = \boldsymbol{\kappa}(\db_e,\db_e) - \k_e^T(\K+\Sig)^{-1}\k_e $        
           		\STATE Report $\hat{\s}_e $, where $p(\hat{\s}_{e}) = \mathcal{N}(\hat{\s}_e|\hat{\mub}_{e},\hat{\boldsymbol{\Sigma}}_{e})$
           \end{algorithmic}
\end{algorithm} 
\vspace{-0.5cm}
\subsection{Approximate linear motion model}\label{linear_model} As an alternative to the general and flexible CGP approach that is applicable to any robot, we also put forth a computationally simple approach that relies on a linear approximation of $\f$. Specifically, if $\Delta t_{jk}$ is sufficiently small, so are elements of $\db_{jk}$. Therefore, it follows from the first order Taylor's series expansion, that $\f$ is approximately linear. This assertion if further verified empirically for the two-wheel differential drive. As evident from Fig. \ref{fig2}, for $\Delta t_{jk}$ sufficiently small, the elements of $\db_{jk}$ are concentrated around zero and the surface fitting them is indeed approximately linear. Motivated by the observation in Fig. \ref{fig2}, we let $\f(\db_{jk}) = \W\db_{jk}$, where $\W \in \Rn^{3 \times m}$ is the unknown weight matrix. The following robust linear regression problem can subsequently be solved to yield the weights:
\begin{align}\label{rlr}
\widehat{\W} = \arg\min_{\W} \sum_{(j,k)\in\E}\sum_{i\in\{x,y,\theta\}} \rho_c\left(\frac{\hat{\s}_{jk}^i - [\W\db_{jk}]_i}{\sigma_{jk}^i}\right) 
\end{align}
where $\rho_c$ is the Huber loss function \cite{huber}. Here, \eqref{rlr} is a convex optimization problem and can be solved efficiently with complexity $\O(n^3)$. It is remarked that the entries of $\W$ do not have any physical significance and cannot generally be related to the intrinsic or extrinsic robot parameters, especially after  wheel deformation. Note that while making predictions the complexity of the linear model is $\mathcal{O}(m)$ where as for CGP it is $\mathcal{O}(n^2)$.

\section{Experimental Evaluations}
This section details the experiments carried out to test the proposed CGP algorithm. We begin with the performance metrics used for validating the accuracy of the estimated model followed by details regarding the experimental setup and results. 

\subsection{Performance Metrics}
In the absence of wheel slippages, it is remarked that the accuracy of estimated model is quantified by the closeness of the robot/sensor trajectory estimate obtained from odometry to the ground truth trajectory. Since ground truth data was not available for the experiments, we instead used a SLAM algorithm to localize the sensor and build a map of the environment. While SLAM output would itself be not as accurate as compared to the ground truth, of which some of them \cite{slam} do not require odometry measurements and consequently serves as a benchmark for all calibration algorithms. Specifically, the \textit{google cartographer} algorithm, which leverages a robust scan to sub-map joining routine, is used for generating the trajectory and the map \cite{slam} of the environments. It is remarked that in the absence of extrinsic calibration parameters, SLAM outputs only the sensor trajectory (and not the robot trajectory), which is subsequently used for comparisons.

Various sensor trajectory estimates are compared on the basis of Relative Pose Error (RPE) and the Absolute Trajectory Error (ATE) motivated from \cite{benchmark}. The RPE measures the local accuracy of the trajectory, and is indicative of the drift in the estimated trajectory as compared to the ground truth. At any time $t_k \in \mathcal{T}$ ,  let the odometry and SLAM pose estimates be denoted by $\hat{\x}_k$ and $\x_k$, respectively. Then, relative pose change between times $t_k$ and $t_{k+1}$ estimated via odometry and SLAM are given by $\circleddash\ \hat{\x}_k \oplus \hat{\x}_{k+1}$ and $\circleddash\ \x_k \oplus \x_{k+1}$, respectively. Defining $\e^r_k := \circleddash\ (\circleddash\ \hat{\x}_k \oplus \hat{\x}_{k+1})\oplus (\circleddash\ \x_k \oplus \x_{k+1})$, the RPE is defined as the root mean square of the translational components of $\{\e^r_k\}_{k=1}^{n-1}$, i.e.,
\begin{equation}
\text{RPE} := \left ( \frac{1}{n-1} \sum\limits_{k=1}^{n-1} \left \|\text{trans}(\e^{r}_k)  \right \|^2 \right )^{1/2}
\end{equation}
where trans$(\textbf{e}_k)$ refers to the translational components of $\textbf{e}_k$. In contrast, the ATE measures the global (in)consistency of the estimated trajectory and is indicative of the absolute distance between the poses estimated by odometry and SLAM at any time $t_k$. Defining the absolute pose error at time $t_k$ as $\e^a_k:=\circleddash\ \hat{\x}_k \oplus \x_k$,  the ATE is evaluated as the root mean square of the pose errors for all times $t_k\in \mathcal{T}$, i.e., 
\begin{equation}
\text{ATE}  := \left ( \frac{1}{n} \sum\limits_{k=1}^{n}  \left \|\text{trans}(\e^a_k)  \right \|^2 \right )^{1/2}.
\end{equation}
Next, we detail the experimental setup used to test model estimates from different forms of Gaussian process (GP).   

\begin{figure}
\begin{subfigure}{.25\textwidth}
  \centering
 \includegraphics[width=0.95\linewidth,trim={0cm 4cm 0cm 0cm},clip]{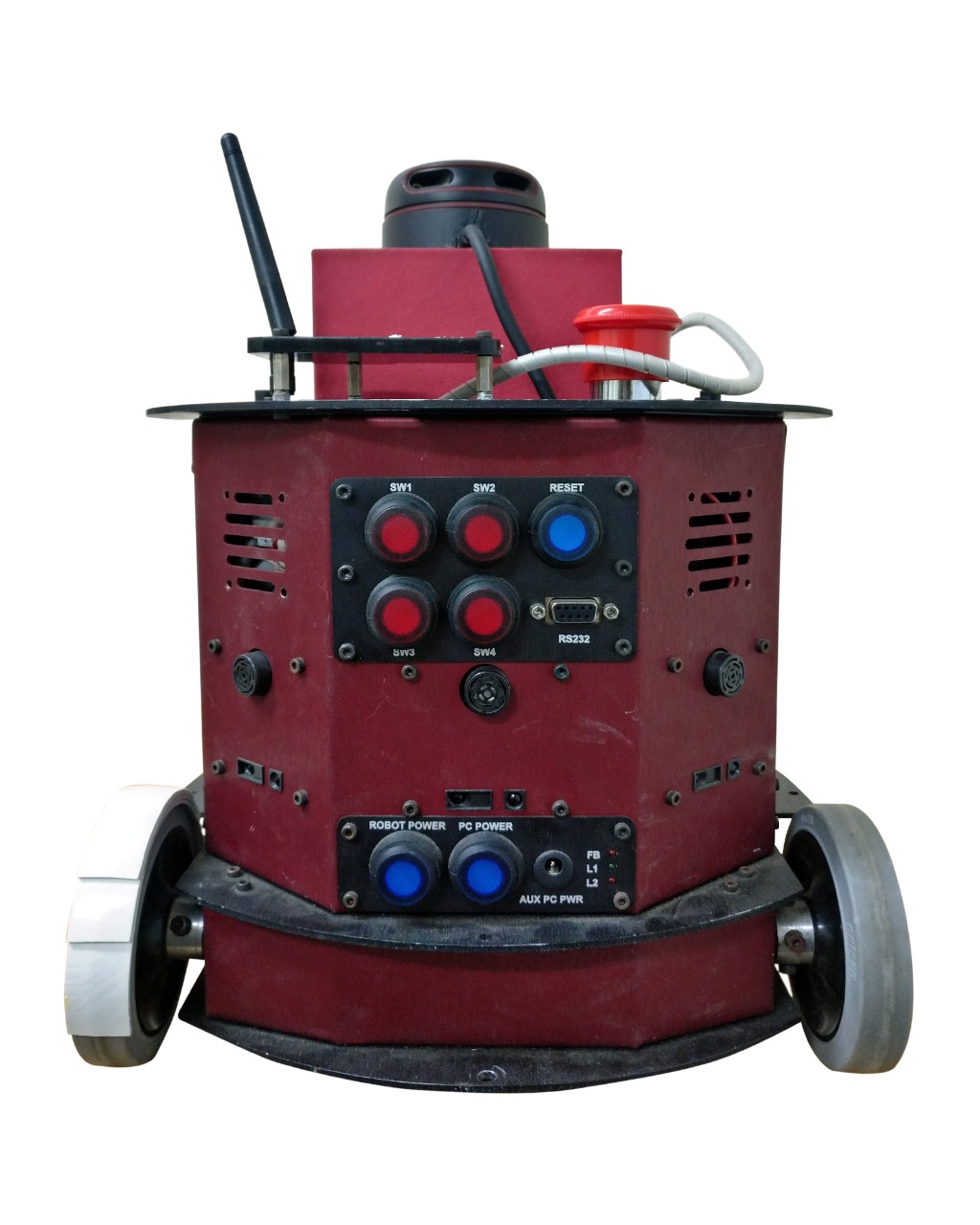}
  \caption{\textit{Fire Bird}} 
\end{subfigure}%
\begin{subfigure}{.25\textwidth}
  \centering
  \includegraphics[width=0.95\linewidth,trim={2cm 1.5cm 2cm 4cm},clip ]{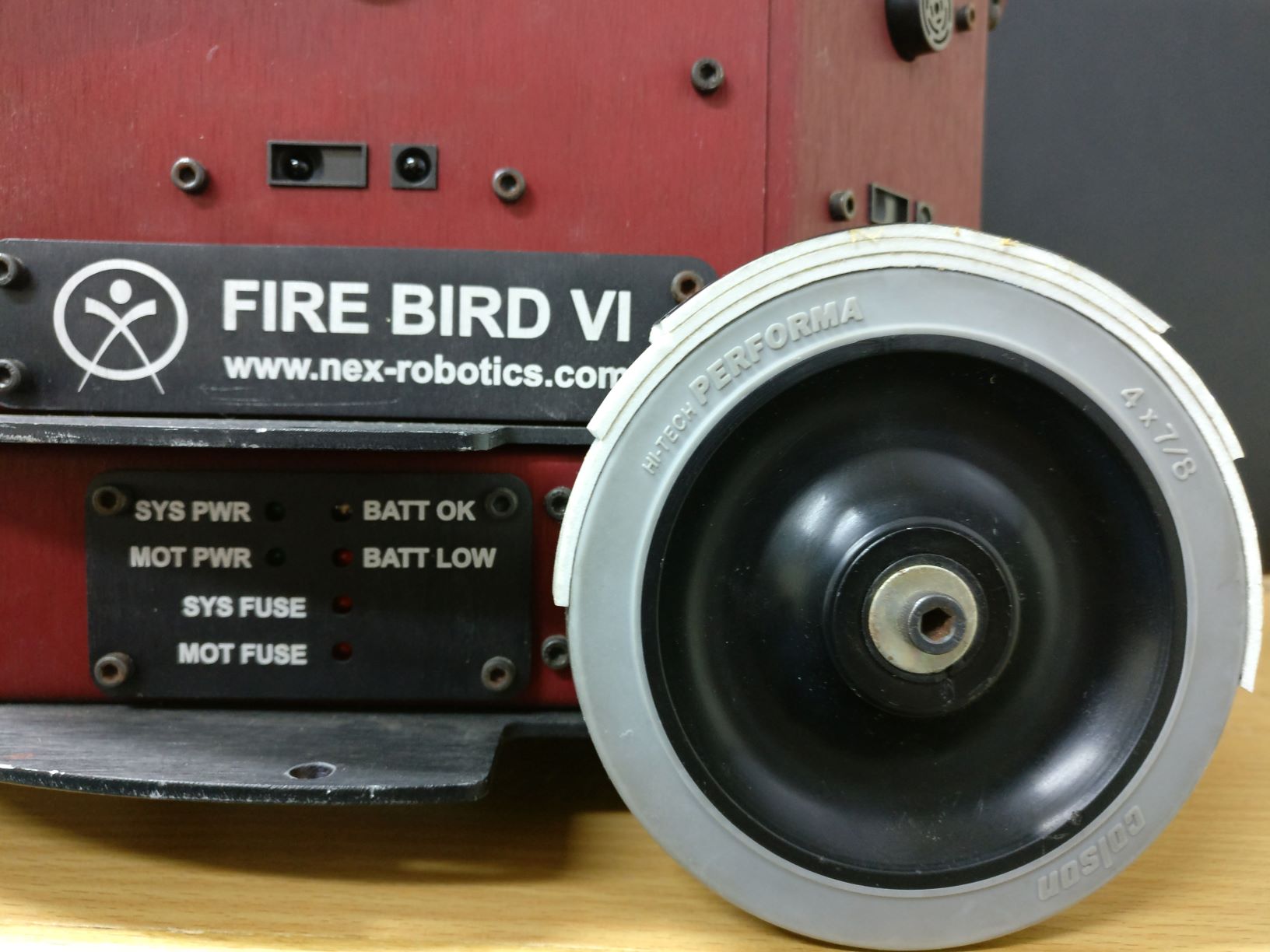}
  \caption{Deformed left wheel} 
\end{subfigure}
\caption{\textit{Fire Bird VI} robot used for experimental evaluations. }\label{fireb}
\end{figure}


\begin{figure*}
\begin{subfigure}{.33\textwidth}
  \centering
 \includegraphics[width=.99\linewidth,trim={0cm 0cm 0cm 0cm},clip ]{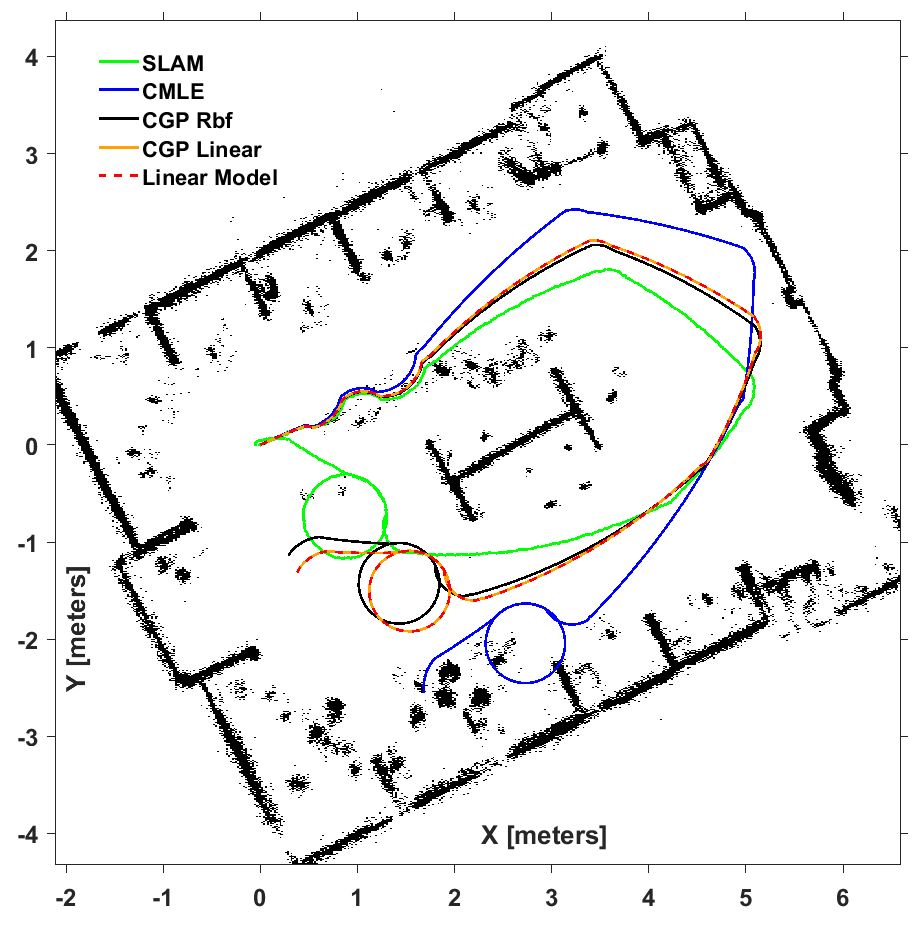}
  \caption{Tomography Lab, Configuratoin \textbf{F1}} 
\end{subfigure}%
\begin{subfigure}{.33\textwidth}
  \centering
 \includegraphics[width=0.99\linewidth,trim={0cm 0cm 0cm 0cm},clip]{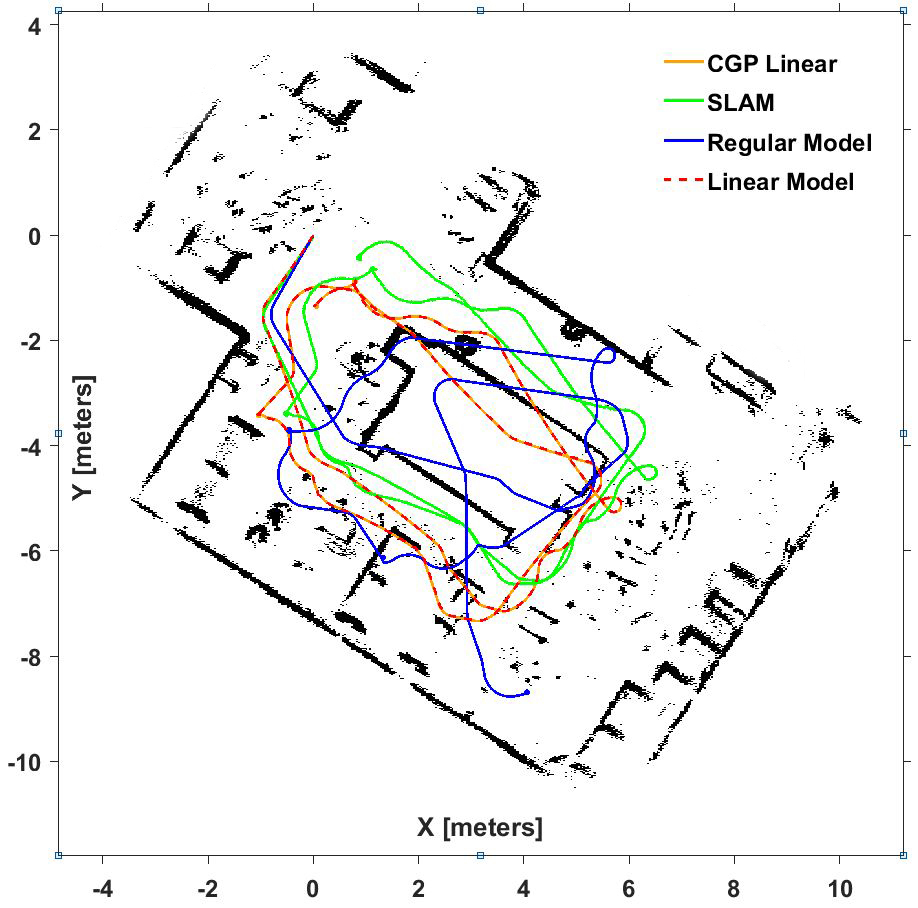}
  \caption{ACES Library, Configuration \textbf{T1}} 
\end{subfigure}
\begin{subfigure}{.33\textwidth}
  \centering
  \includegraphics[width=0.99\linewidth,trim={0cm 0cm 0cm 0cm},clip]{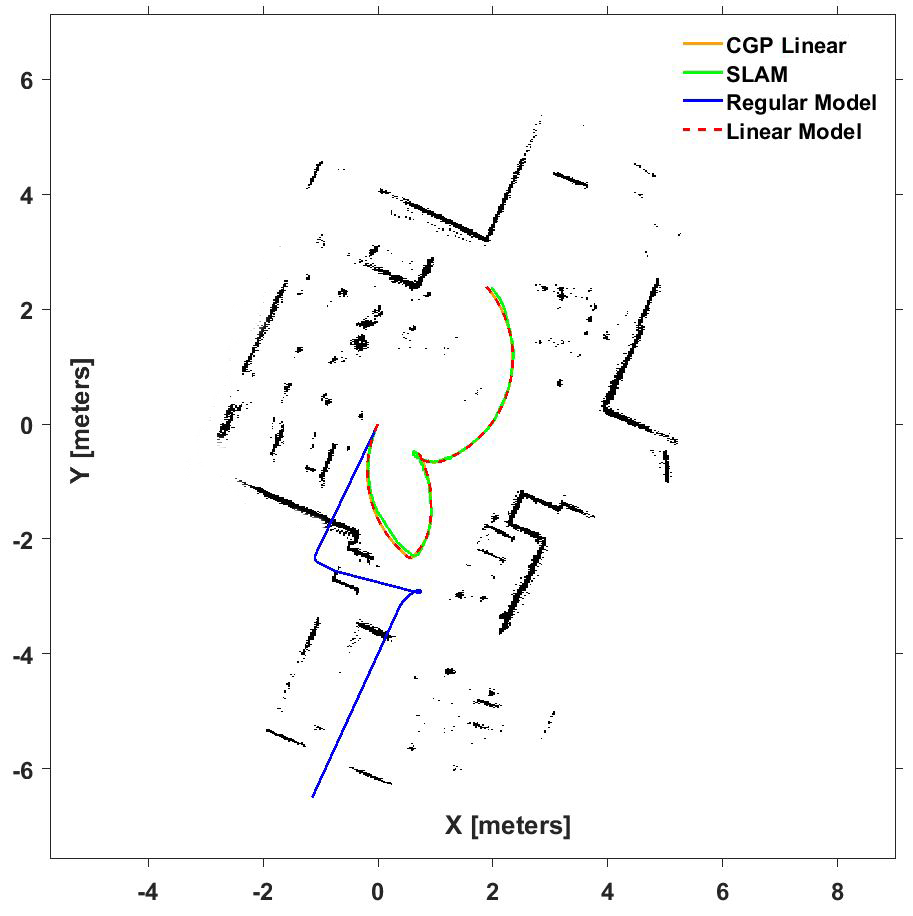}
  \caption{ACES Library, Configuration \textbf{T2}} 
\end{subfigure}
\caption{Trajectory comparison against SLAM for different robot configurations. (a) is the test environment for the configuration \textbf{F1} where as (b) and (c) are for configurations \textbf{T1} and \textbf{T2} respectively.}
 	\vspace{-0.5cm} \label{traj_all} 
\end{figure*}

\begin{table}\fontsize{8pt}{20pt}\selectfont
\centering
\caption{List of experimental configurations with labels and locations}
\label{config}
\begin{tabular}{|c|c|c|c|}
\hline
\textbf{Robot} & \textbf{Configuration} & \textbf{\begin{tabular}[c]{@{}c@{}}Training \\ Data\end{tabular}} & \textbf{Test Data} \\ \hline
\textit{FireBird VI} & \textbf{F1} & WSN Lab & \begin{tabular}[c]{@{}c@{}}Tomography\\  Lab\end{tabular} \\ \hline
\multirow{2}{*}{\textit{Turtlebot3}} & \textbf{T1} & \multirow{2}{*}{\begin{tabular}[c]{@{}c@{}}ACES \\ Library\end{tabular}} & \multirow{2}{*}{\begin{tabular}[c]{@{}c@{}}ACES \\ Library\end{tabular}} \\ \cline{2-2}
 & \textbf{T2} &  &  \\ \hline
\end{tabular}
\end{table}

\begin{table}\fontsize{8pt}{14pt}\selectfont
\centering
	\caption{ ATE and RPE for Configurations \textbf{F1}}
	\label{2wheel}
\begin{tabular}{|c|c|c|c|}
\hline
\multicolumn{2}{|c|}{\textbf{GP Estimate}} & \multicolumn{2}{c|}{\textbf{Configuratin F1}} \\ \hline
\textbf{Mean fn} & \textbf{Kernel fn} & \textbf{ATE (m)} & \textbf{RPE (mm)} \\ \hline
Zero & RBF & 6.273 & 9.634 \\ \hline
Linear & RBF & \textbf{0.592} & \textbf{9.367} \\ \hline
Zero & Linear & {0.687} & {9.367} \\ \hline
Zero & RBF + Linear & 0.716 & 9.34 \\ \hline
Linear & RBF + Linear & 0.732 & 9.343 \\ \hline \hline
\multicolumn{2}{|c|}{Linear Model} & \textbf{0.687} & \textbf{9.367} \\ \hline
\multicolumn{2}{|c|}{CMLE \cite{censi}} & 1.546 & 9.361 \\ \hline
\end{tabular} 
\end{table}

\begin{table}\fontsize{12pt}{20pt}\selectfont
	\caption{ ATE and RPE for Configurations \textbf{T1} and \textbf{T2}}
	\label{erro_tur}
	\resizebox{\columnwidth}{!}{%
\begin{tabular}{|c|c|c|c|c|c|}
\hline
\multicolumn{2}{|c|}{\textbf{GP Estimate}} & \multicolumn{2}{c|}{\textbf{Configuration \textbf{T2}}} & \multicolumn{2}{c|}{\textbf{Configuration \textbf{T1}}} \\ \hline
\textbf{Mean fn} & \textbf{Kernel fn} & \textbf{ATE (m)} & \textbf{RPE (mm)} & \textbf{ATE (m)} & \textbf{RPE (mm)} \\ \hline
Zero & RBF & 2.49 & 5.21 & 6.523 & 5.466 \\ \hline
Linear & RBF & 0.161 & 8.324 & 5.006 & 5.573 \\ \hline
Zero & Linear & \textbf{0.068} & \textbf{5.108} & \textbf{0.87} & \textbf{5.457} \\ \hline
Zero & RBF + Linear & 3.798 & 5.121 & 0.87 & 5.455 \\ \hline
Linear & RBF + Linear & 1.193 & 5.49 & 1.849 & 5.519 \\ \hline \hline
\multicolumn{2}{|c|}{Linear Model} & \textbf{0.068} & \textbf{5.108} & \textbf{0.869} & \textbf{5.458} \\ \hline
\multicolumn{2}{|c|}{Regular Model} & 4.24 & 5.112 & 3.647 & 5.517 \\ \hline
\end{tabular}}
\end{table}

\subsection{Experimental Setup}
\subsubsection{Robots}  We have used a two-wheel differential drive  \emph{FireBird VI} robot (see Fig. \ref{fireb}) having a particular set of intrinsic parameters \cite{firebird_web} and a four-wheel mecanum drive  \emph{Turtlebot3} robot (see Fig. \ref{meca}). The \emph{Fire Bird VI} is primarily a research robot with diameter 280 mm, weight of 12 kilograms, and maximum translational velocity of 1.28 m/s. All \textit{FireBird} encoders publish data at the rate of 10 Hz with a resolution of 3840 ticks per revolution. Similarly \emph{Turtlebot3 mecanum} is also a research robot from the Robotis group with all wheels diameter of 60 mm. It weighs 1.8 kilograms and maximum translational velocity is 0.26m/s. The dynamixels used publish data at 10 Hz with an approximate resolution of 4096 ticks per revolution. For the purposes of the experiment, we made use of an on board computer with i5 processor, 8GB RAM, running ROS kinetic for processing the data from lidar and wheel encoders, performing SLAM for validation, and running the calibration algorithms. 
 
\subsubsection{Lidar Sensor} RPLidar A2 is a low cost $360^\circ$, $2D$ laser scanner with a detection range of 6 meters, a distance resolution less than 0.5 m and an adjustable operating frequency of 5 to 15 Hz. This scanner was mounted on the both the robots with the frequency of 10 Hz resulting in an angular resolution of 0.9$^\circ$.

\subsubsection{Scan Matching} We used point-to-line ICP (PLICP) variant \cite{P2LICP} in order to estimate the sensor displacements $\hat{\s}_{jk}$. It is remarked that all ICP-like methods also output the corresponding covariance value in closed-form \cite{covariance} that can be used by the CGP algorithm.
  
\subsubsection{Data Processing} For the purposes of the experiments, we ensure that scans are collected at times spaced $T$ seconds apart. The choice of $T$ is not trivial. For instance, choosing a small $T$ often makes the algorithm too sensitive to un-modeled effects arising due to synchronization of sensors, robot's dynamics. It is lucrative to choose far scan pairs as more information is capured about the parameters however both the scan matching output as well as the motion model become inaccurate when $T$ is large. For the experiments, we chose the largest value of $T$ that yielded a reliable scan matching output in the form of sensor motion, resulting in $T=0.6$ second for \emph{Turtlebot3 mecanum} and $T = 0.3$ seconds for the \emph{Firebird VI} robot. These values are chosen based on maximum wheel speeds such that slippages are minimized during experimentation. Note that since the odometry readings are acquired at a rate, higher than the scans, temporally closest odometry reading is associated to a given scan. With the chosen $T$ the robots would move a maximum displacement of 15cm in x and y and 8$^\circ$ in yaw, under such conditions PLICP achieves 99.51 $\%$ accuracy \cite{P2LICP}.

\subsubsection{Deformed Robot Configurations}  In order to demonstrate the non-availability of the robot model, one of the wheels of the \emph{Firebird VI} robot is deformed with a thick tape (see Fig. \ref{fireb}(b)), this configuration is referred as \textbf{F1}. Care was taken to ensure that the deformation was not too large, so as to avoid wobbling of the robot and the scan plane of the Lidar. In the case of  \emph{Turtlebot3} robot two different configurations (\textbf{T1} and \textbf{T2}) are constructed (see Fig. \ref{meca}), by changing the position of the wheels from the regular configuration. We will see further that the amount of deformation in tilted wheel configuration \textbf{T2} (as in Fig. \ref{meca}(b)) is more as opposed to unaligned wheels configuration \textbf{T1} (as in Fig. \ref{meca}(a)). Next, experiments comprising of training and testing phases, are carried out using these deformed robots for all the specified configurations (see Table \ref{config}). While training data is used to learn the motion model of the robot/sensor, the test data is used to evaluate the accuracy of the learned model. It is remarked that the collected test data involves short and long trajectories with varied robot motions. Each experiment is labeled for reference, with details provided as shown in Table \ref{config}. For example, configuration \textbf{T1} refers to the experiment done using \emph{Turtlebot3} robot, where both training and test data are collected in ACES library.
 
 \subsection{Experimental Results}\label{expres}  
We first perform offline calibration of \textit{FireBird VI} robot with configuration \textbf{F1} using the proposed CGP algorithm along with the model based CMLE \cite{censi} algorithm. Note that in the case of CGP algorithm various kernel and mean functions are trained to determine which of them captures the sensor motion model accurately. Note that we have also trained on composite kernel functions like $\kappa_{rbf} + \kappa_{lin} (\textbf{x},\textbf{x}')$. After the model is learned, predictions are made on the test data. The predicted trajectories are then compared with SLAM trajectory as reference. Error metrics for these trajectories are generated and displayed in Table \ref{2wheel}. It is observed that CGP with squared exponential kernel function with linear mean function outperforms other trained models, also CMLE. We remark here that although CMLE predicts the radius of the left wheel to be slightly more than that of the
right wheel, the predictions are worse due to non applicability of the parametric model as the wheel looses its notion of circularity. Observe that CGP with linear kernel function is comparable to the best case.  
The predicted trajectories generated for CMLE \cite{censi}, CGP with linear kernel and SLAM \cite{slam} are displayed in Fig. \ref{traj_all}(a). It is evident that the proposed CGP with linear kernel predicts test trajectory close to SLAM. 

Similar procedure is carried out in the case of Turtlebot3 robot with configurations \textbf{T1} and \textbf{T2}. Note that since the analysis of CMLE \cite{censi} is restricted to two wheel differential drive robots, we use parametric motion model of four wheel mecanum drive robot \cite{mecanum} with manufacturer specified parameters for robot intrinsics and nominal hand measured parameters for lidar extrinsics to perform predictions on test data. Table \ref{erro_tur} displays error metrics evaluated for parametric and various non-parametric models. Observe that the proposed CGP algorithm with linear kernel function outperforms other learned models. The corresponding test trajectories for configurations \textbf{T1} and \textbf{T2} are displayed in Fig. \ref{traj_all}(b) and Fig. \ref{traj_all}(c) respectively.

Interestingly it can be observed from Table.\ref{2wheel} and Table \ref{erro_tur} that the linear model approximation is sufficient to explain the motion model with the set deformities in all configurations. Here we notice that learning a linear approximation of $\f$ is sufficient to accurately predict robot odometry, this is in lines with our discussion in Sec. \ref{linear_model}.

 \section{Conclusion}
We develop a novel odometry and sensor calibration framework applicable to wheeled mobile robots operating in planar environments. The key idea is to utilize the ego-motion estimates from the exteroceptive sensor to estimate the motion model of the sensor/robot. The proposed framework is general as it applies to robots whose motion model is not known. We advocate a non-parametric Gaussian process regression-based approach that directly learns the relationship between the wheel odometry and the sensor motion. The method does not require ground-truth measurements from an external setup, and henceforth the calibration routine can be carried out without interrupting the robot operation. A computationally efficient method that relies on a linear approximation of the sensor motion model is shown to perform on par with the proposed calibration via Gaussian process (CGP) algorithm. Experiments are performed on robots with un-modelled deformations and is shown to outperform existing parametric approaches. Moreover, all the MATLAB codes are made available online\footnote{http://www.tinyurl.com/GPcalibration}. The method being general is applied to wheeled robots operating in planar environments but does not make any assumptions regarding the same. As part of the future work, it would be interesting to test the performance in non-planar settings.

 \footnotesize
 \bibliographystyle{IEEEtran}
 \bibliography{refer}

\end{document}